\documentclass[12pt]{article}

\usepackage[english]{babel}
\usepackage{amssymb}
\usepackage{a4wide}
\usepackage{psfig}

\newcommand{\be}{\begin{equation}}
\newcommand{\ee}{\end{equation}}

\title{{\Large {\bf
Intelligence and Cooperative Search by Coupled Local Minimizers
}}
\author{{\bf Johan A.K. Suykens,} {\bf Joos Vandewalle} 
{\bf and Bart De Moor}\\ \\
Katholieke Universiteit Leuven \\
Department of Electrical Engineering, ESAT-SISTA \\
Kardinaal Mercierlaan 94, B-3001 Leuven (Heverlee), Belgium \\
Tel: 32/16/32 18 02 \, \, Fax: 32/16/32 19 70 \\
Email: johan.suykens@esat.kuleuven.ac.be \\ \\
Corresponding author: Johan Suykens \\ \\
Running title: Intelligence and Cooperative Search 
}}
\date{{\bf Published in} \\
{\bf Int. J. Bifurcation and Chaos, Vol.11, No.8, pp.2133-2144, 2001}}

\begin{document}
\maketitle

\newpage


\begin{abstract}
We show how coupling of local optimization processes can lead to better 
solutions than multi-start local optimization consisting of independent runs.
This is achieved by minimizing the average energy cost of the ensemble, 
subject to synchronization constraints between the state 
vectors of the individual local minimizers. From an augmented Lagrangian
which incorporates the synchronization constraints both as soft and
hard constraints, a network is derived wherein the local minimizers
interact and exchange information through the synchronization constraints.
From the viewpoint of neural networks, the array can be considered as a 
Lagrange programming network for continuous optimization
and as a cellular neural network (CNN).
The penalty weights associated with the soft state synchronization
constraints follow from the solution to a linear program.
This expresses that the energy cost of the ensemble should maximally decrease. 
In this way successful local minimizers can implicitly impose
their state to the others through a mechanism of master-slave 
dynamics resulting into a cooperative search mechanism.
Improved information spreading within the ensemble is obtained
by applying the concept of small-world networks.
We illustrate the new optimization method on two different problems:  
supervised learning of multilayer perceptrons
and optimization of Lennard-Jones clusters. 
The initial distribution of the local minimizers
plays an important role. For the training of multilayer perceptrons this 
is related to the choice of the prior on the interconnection weights 
in Bayesian learning methods. 
Depending on the choice of this initial distribution, coupled local
minimizers (CLM) can avoid overfitting and produce good generalization,
i.e.~reach a state of intelligence.
In potential energy surface optimization of Lennard-Jones clusters,
this choice is equally important. In this case it can be 
related to considering a confining potential.
This work suggests, in an interdisciplinary context,
the importance of information exchange 
and state synchronization within ensembles,  
towards issues as evolution, collective behaviour, optimality and 
intelligence. 
\end{abstract}

\newpage

\section{Introduction}

A large variety of problems arising in engineering, physics, chemistry and 
economics can be formulated as optimization problems, either
constrained or unconstrained having continuous or discrete variables.
A well developed area in optimization theory are local optimization methods
including conjugate gradient, quasi-Newton,
Levenberg-Marquardt, sequential quadratic programming etc. 
[Bertsekas, 1996; Fletcher, 1987; Gill {\em et al.}, 1981].
Continuous-time optimization methods have been developed
within the area of neural networks [Cichocki \& Unbehauen, 1994].
Popular methods for global exploration of the search space
are e.g.~simulated annealing [Kirkpatrick {\em et al.}, 1983] and 
genetic algorithms [Goldberg, 1989].
Within standard local optimization techniques one often applies
multi-start local optimization, by trying different starting points
and running the processes independently from each other and selecting
the best result from all trials. 
On the other hand, for training of neural networks such as 
multilayer perceptrons (MLP) 
it is well-known that instead of training several MLPs for random choices of 
small initial weights and selecting the best of all trained networks, one
better forms a committee network which is based upon all trained networks
[Arbib, 1995; Bishop, 1995].
In other words, the training efforts of the less optimal networks
are not entirely useless but can be employed in order to improve the estimate
in view of the bias-variance trade-off [Bishop, 1995]. 
Unfortunately, committee networks are only applicable
to static nonlinear regression and classification problems.

In this paper we propose a new methodology of coupled local minimizers (CLM)
for solving continuous nonlinear optimization problems.
We pose a somewhat similar challenge as for committee networks
but within a different and broader context of solving 
differentiable optimization problems. The aim is   
to (on-line) combine the results from local optimizers in order 
to let the ensemble generate a local minimum that is better than the best
result obtained from all individual local minimizers.
We show how improved local minima can be obtained by having interaction
and information exchange between the local search processes.
This is realized through state synchronization constraints that are 
imposed between the local minimizers by incorporating principles
of master-slave dynamics.
Synchronization theory has been intensively studied within the area of 
chaotic systems and secure communications
[Chen \& Dong, 1998; Pecora \& Carroll, 1990;
Suykens {\em et al.}, 1996, 1997, 1998; Wu \& Chua, 1994]. 
The CLM method is related to Lagrange programming
network approaches for chaos synchronization [Suykens \& Vandewalle, 2000],
where identical or generalized synchronization constraints are imposed 
on dynamical systems.
CLMs also fit within the framework of Cellular 
Neural Networks (CNN) [Chua \& Roska, 1993; Chua {\em et al.}, 1995; Chua, 1998].
By considering the objective of minimizing the average cost of 
an ensemble of local minimizers subject to pairwise synchronization
constraints, a continuous-time optimization algorithm
is studied according to Lagrange programming networks
[Cichocki \& Unbehauen, 1994; Zhang \& Constantinides, 1992].
The resulting continuous-time optimization algorithm
is described by an array of coupled nonlinear cells or a 
one-dimensional CNN with bi-directional coupling.

We show how to obtain intelligence from CLMs. 
This is done by considering a problem of static nonlinear regression
with MLPs from given noisy measurement data.
In this case CLMs correspond
to coupled backpropagation processes [Rumelhart {\em et al.}, 1986; Werbos, 1990].
In order to obtain an MLP with good generalization performance 
(i.e.~an intelligent solution) it is usually needed to consider 
a regularization term in addition to the original sum squared error
cost function (fitting error) which is defined on the training data
[Bishop, 1995; MacKay, 1992; Poggio \& Shelton C. 1999; 
Suykens \& Vandewalle, 1998]. When applying CLMs such a 
regularization term is not needed. It turns out that 
the initial distribution of the initial states of the several 
local minimizers can play this role, such that one only has to 
optimize the training set error. This is consistent with 
insights from Bayesian learning of neural networks where 
the regularization term is related to the choice of the prior
on the unknown interconnection weights, which expresses that 
the initial weights should be chosen small (weight decay).

The role of the initial CLM state is not only important for 
neural networks but also with respect to 
potential energy surface optimization of Lennard-Jones 
clusters, a second class of problems that is investigated in
this paper. It is considered to be an important benchmark problem 
in the area of protein folding
[\v{S}ali {\em et al.}, 1994; Neumaier, 1997; Wales \& Scheraga, 1999].
We show that CLMs are able to detect the global minimum
of (LJ)$_{38}$ which possesses a {\em double-funnel} energy landscape and is known
to be a challenging test-case
[Wales \& Doye, 1997; Wales {\em et al.}, 1998; Wales \& Scheraga, 1999].
The CLM method has also been successfully applied 
to larger scale problems. The cooperative search mechanism in CLMs
is obtained by solving a linear programming problem in the unknown
soft constraint penalty factors. In this way a maximal decrease
in the average energy cost of the ensemble is achieved.

This paper is organized as follows. In Section 2 we introduce
coupled local minimizers. In Section 3 we discuss how to obtain
optimal interaction between local minimizers. In Section 4 
we present new insights on coupled backpropagation processes
and intelligence. In Section 5 we apply coupled local minimizers
to the optimization of Lennard-Jones clusters.


\section{A CNN of Coupled Local Minimizers}

Consider the minimization of a twice continuously differentiable 
cost function $U({\bf x})$ with ${\bf x} \in {\Bbb R}^{n}$.
Let us take an ensemble consisting of $q$ local minimizers.
We aim at minimizing the average energy cost
$\langle U \rangle = \frac{1}{q} \sum_{i=1}^{q} U[{\bf x}^{(i)}]$ 
of this ensemble subject to pairwise state synchronization of the particles:
\be
\label{costfu}
\min_{{\bf x}^{(i)} \in {\Bbb R}^{n}} \, \langle U \rangle \,\, 
{\rm such \,\, that} \,\, {\bf x}^{(i)} - {\bf x}^{(i+1)} 
= 0 , \,\, \,\, \, i=1,2,...,q
\ee
with particle states ${\bf x}^{(i)} \in {\Bbb R}^{n}$ for $i=1,2,...,q$ and
boundary conditions ${\bf x}^{(0)} = {\bf x}^{(q)}$, ${\bf x}^{(q+1)} = {\bf x}^{(1)}$.
The synchronization constraints have to be achieved in an 
asymptotical sense, i.e.~the particles have to reach the same final state. 
One defines the augmented Lagrangian
\be
\label{lagrangian}
{\cal L}({\bf x}^{(i)}, {\bf \lambda}^{(i)}) = 
\frac{\eta}{q} \sum_{i=1}^{q} \, U[{\bf x}^{(i)}] +
\frac{1}{2} \sum_{i=1}^{q} \, \gamma_{i} \, \| {\bf x}^{(i)} - 
{\bf x}^{(i+1)} \|_{2}^{2} +
\sum_{i=1}^{q} \langle {{\bf \lambda}^{(i)}} , [{\bf x}^{(i)} - {\bf x}^{(i+1)}] \rangle
\ee
with Lagrange multipliers ${\bf \lambda}^{(i)} \in {\Bbb R}^{n}$ $(i=1,2,...,q)$.
The different terms in this Lagrangian are the objective function,
the soft constraints and the hard constraint related to the pairwise 
state synchronization constraints. The penalty factors $\gamma_{i}$
emphasize the importance of each of the soft synchronization constraints.
From this augmented Lagrangian one obtains the Lagrange programming network 
[Zhang \& Constantinides, 1992]
\be
\label{LPN}
\left\{
\begin{array}{rcl}
\dot{{\bf x}}^{(i)} & = &  - \nabla_{{\bf x}^{(i)}}^{ }
{\cal L}({\bf x}^{(i)}, {\bf \lambda}^{(i)})  \\
\dot{{\bf \lambda}}^{(i)} & = & 
\nabla_{{\bf \lambda}^{(i)}}^{ } {\cal L}({\bf x}^{(i)}, {\bf \lambda}^{(i)}) \, , \,
i=1,2,...,q.
\end{array}
\right.
\ee
This is basically a continuous-time optimization algorithm for 
solving the given constrained optimization problem.
One obtains the following array of coupled local minimizers (CLM) 
\be
\label{CLM}
\left\{
\begin{array}{rcl}
\dot{{\bf x}}^{(i)} & = &  - \frac{\eta}{q} \nabla_{{\bf x}^{(i)}} U[{\bf x}^{(i)}] +
\gamma_{i-1} [{\bf x}^{(i-1)} - {\bf x}^{(i)}] -
\gamma_{i} [{\bf x}^{(i)} - {\bf x}^{(i+1)}] + {\bf \lambda}^{(i-1)} - 
{\bf \lambda}^{(i)} \\ 
\dot{{\bf \lambda}}^{(i)} & = & {\bf x}^{(i)} - {\bf x}^{(i+1)} \, , \,
i=1,2,...,q 
\end{array}
\right.
\ee
with learning rate $\eta$. This array consists of a number of 
$q$ coupled nonlinear cells and is considered to be a special case of 
cellular neural networks (CNN) [Chua, 1998].

The basic mechanism of state synchronization with coupled 
local minimizers is illustrated in Fig.~1 for a double potential
well problem with two particles that are searching for the global
minimum. The main idea here is to impose that they should reach
the same final position. A simplified CLM considered in this case is
\be
\label{CLM_doublewell}
\left\{
\begin{array}{rcl}
\dot{x} & = & - \nabla_{x}^{ } U(x) - (x-z) - \lambda \\
\dot{z} & = & - \nabla_{z}^{ } U(z) + (x-z) + \lambda \\
\dot{\lambda} & = & x - z
\end{array}
\right.
\ee
where $x, z, \lambda \in {\Bbb R}$. This is
derived from the augmented Lagrangian
\be
\label{lag_doublewell}
{\cal L}(x, z, \lambda) = 
\frac{\eta}{2} [U(x) + U(z)] +
\frac{1}{2} (x - z)^{2} + \lambda (x - z) .
\ee
As a result of the constraining of 
the system, the two particles are enforced to take a decision
about which valley to choose. When the initial states of the 
two particles are located in different valleys, one observes
that they are always reaching the global minimum.
This phenomenon is independent from the steepness and shape
of the valleys. In this process, the search space is in fact duplicated
and the local optimizers are exchanging information through
the synchronization constraint. 

Note that it is not necessarily guaranteed beforehand that the equality 
constraints in (\ref{costfu}) will be {\em exactly} realized by the CLM, as is known
for Lagrange programming networks. In this case one could 
consider as CLM goal that the state synchronization 
constraints should be realized in an approximate sense, 
i.e.~bringing the states {\em close} to each other. Eventually, the equality 
constraints could be replaced by a set of inequality constraints which
would implement this.


\section{CLMs and Optimal Cooperative Search}

In this Section, we discuss how to obtain an optimal interaction 
between the several local minimizers. This is done by making
a suitable choice of the penalty weights $\gamma_{i}$.
In this design procedure we aim at maximally decreasing the 
instantaneous average energy cost of the ensemble
\be
\label{selection}
\begin{array}{rcl}
{\displaystyle \frac{d \langle U \rangle }{dt}} & = & {\displaystyle 
\frac{1}{q} \sum_{i=1}^{q} 
\langle \frac{\partial U[{\bf x}^{(i)}]}{\partial {\bf x}^{(i)}} ,
\dot{{\bf x}}^{(i)} \rangle } \\ \vspace*{-3mm} \\
 & = &
{\displaystyle \frac{1}{q} 
\sum_{i=1}^{q}  \langle \frac{\partial U[{\bf x}^{(i)}]}{\partial {\bf x}^{(i)}} ,
 - \frac{\eta}{q} \frac{\partial U[{\bf x}^{(i)}]}{\partial {\bf x}^{(i)}} +
\gamma_{i-1} [{\bf x}^{(i-1)} - {\bf x}^{(i)}] } \\ & & 
{\displaystyle - \gamma_{i} [{\bf x}^{(i)} - {\bf x}^{(i+1)}] + 
\lambda^{(i-1)} - \lambda^{(i)} \rangle } .
\end{array}
\ee
For a given state vector ${\bf x} = [{\bf x}^{(1)}; {\bf x}^{(2)}; ...; {\bf x}^{(q)}]$
and costate of the ensemble 
${\bf \lambda} = [{\bf \lambda}^{(1)}; \lambda^{(2)}; ...;$ $ \lambda^{(q)}]$
the latter expression is affine in 
${\bf \gamma} = [\gamma_{1}; \gamma_{2}; ... ; \gamma_{q}]$
and the interaction between the local minimizers is
optimized then by solving the linear program (LP):
\be
\min_{{\bf \gamma} \in {\Bbb R}^{q}} \, 
\frac{d \langle U \rangle}{dt}\mid_{{\bf x}, {\bf \lambda}}
\,\, \,\, \,\, {\rm such \,\, that} \,\, \,\, \,\,
\underline{\gamma} < \gamma_{i} < \overline{\gamma} , \,\, \,\,\, i=1,2,...,q
\ee
where $\underline{\gamma}, \overline{\gamma}$ are user-defined
lower- and upperbounds. The resulting $\gamma_{i}$ values are kept constant
for simulation of the CLM over a certain time interval $\Delta T$.
The difference between the $\gamma_{i}$ values in (\ref{CLM})
causes a similar effect as is known in master-slave 
(drive-response) synchronization which has been studied for 
secure communications using chaos [Chen \& Dong, 1998; Pecora \& Carroll, 1990;
Suykens {\em et al.}, 1997, 1998; Wu \& Chua, 1994].
In this way, successfully performing local optimizers impose their state
to the other part of the ensemble, which leads to cooperative behaviour.

The step size $\eta$ is determined here by imposing a target evolution law
\be
d( \langle U \rangle - U^{*})/dt = - \alpha \, ( \langle U \rangle - U^{*}) 
\ee
where $\alpha$ is a user-defined positive real constant
and $U^{*}$ is an estimate of the energy cost value at the global minimum
(alternatively one adds a constant value to the cost function
such that the energy cost is guaranteed to be positive everywhere).
As a result the step size $\eta$ becomes
$\eta = 
[q \sum_{i=1}^{q} \langle \frac{\partial 
U[{\bf x}^{(i)}]}{\partial {\bf x}^{(i)}} , {\bf h} \rangle
+ q \alpha ( \sum_{i} U[{\bf x}^{(i)}] - q U^{*}) ]
/
\sum_{i=1}^{q} \langle \frac{\partial U[{\bf x}^{(i)}]}{\partial {\bf x}^{(i)}} , 
\frac{\partial U[{\bf x}^{(i)}]}{\partial {\bf x}^{(i)}} \rangle $
subject to additional user-defined constraints 
$\underline{\eta} < \eta < \overline{\eta}$ where
${\bf h} = \gamma_{i-1} [{\bf x}^{(i-1)} - {\bf x}^{(i)}] -
\gamma_{i} [{\bf x}^{(i)} - {\bf x}^{(i+1)}] + \lambda^{(i-1)} - \lambda^{(i)}$ 
by definition.
The values of ${\bf \gamma}_{i}, \eta$ are re-scheduled each time in between
simulations of the array over user-defined time intervals $\Delta T$.

According to the idea of small-world networks [Watts \& Strogatz, 1998], the information
spreading within the ensemble can be further improved, e.g.~by
re-numbering a part of the ensemble 
(state vectors together with their corresponding co-state) 
at random every $c \Delta T$ steps 
where $c \in {\Bbb N}_{0}$. In this way the pairwise synchronization
constraints vary among the different local minimizers during optimization. 

In order to obtain insight in the influence of the CLM tuning parameters 
let us consider the optimization of the cost function
\be
\label{example}
\begin{array}{c}
{\displaystyle U({\bf x}) = \frac{1}{2 n} a \sum_{i=1}^{n} 
x_{i}^{2} + 8 n - 4 n \prod_{i=1}^{n} \cos(\omega_{1} x_{i})
- 4 n \prod_{i=1}^{n} \cos(\omega_{2} x_{i}) }
\end{array}
\ee
with $a=0.01$, $\omega_{1} = 0.2$, $\omega_{2} = 1$ in a 
$n=10$ dimensional search space [Styblinski \& Tang, 1990]. 
The global minimum for this problem
is known and located at the origin with $U(0) =0$ and $U^{*} = 0$ is taken. 
The solutions from CLMs improve with respect to multi-start
local optimization, even when a quasi-Newton method instead of
steepest descent optimizers (as for the CLM) are taken.
In these comparisons the initial states have been random 
uniformly generated in $[-20 \,\, 20]^{n}$ 
where the same initial states have been taken for the different 
algorithms and several runs of the algorithms have been done. 
CLMs with $q \geq 20$ are reaching the global minimum in this case. 
From this example it follows that 
in order to optimize a surface over a certain region
in search space, the number $q$ needed to achieve a 
good performance depends on the complex shape of the surface
or typically on the number of local minima per volume in 
search space that one intends to explore. 
Experiments suggest that a factor 
$\overline{\gamma}/\underline{\gamma} = 10$ is usually a good choice.
In this way the energy level or cost function of the optimizers
remains {\em bundled}. Otherwise the simulations could lead to excessive exploration
and decreased speed of convergence.
Typically, larger values of $\overline{\gamma}$ will require shorter 
$\Delta T$ intervals. The choice of the pressure coefficient $\alpha$ 
will control the average energy decrease of the ensemble.
Demanding a high pressure $\alpha$, however, might lead to less exploration
and worse local minima. The bounds on $\eta$  
are taken here as $\underline{\eta} = 10^{-2}$, $\overline{\eta} = 10^{3}$.
Although it is not necessary, 
the CLM performance can be further enhanced 
by applying the small-world network concept, by re-numbering at random
(here 20\% of the local minimizers) 
every $c \Delta T$ steps where $c=5$ typically, leading to improved 
information spreading among the optimizers. For the simulations here and
in the sequel a Runge-Kutta integration rule with adaptive step size 
(absolute and relative error equal to $10^{-2}$) has been used.


\section{Intelligence and Coupled Backpropagation}

In this Section we apply CLMs to the training of 
multilayer perceptron networks
\be
\label{MLP}
y = w^{T} \tanh(V u + \beta) 
\ee
with input vector $u \in {\Bbb R}^{m}$, output $y \in {\Bbb R}$,
interconnection weights $w \in {\Bbb R}^{n_{h}}$, $V \in {\Bbb R}^{n_{h} \times m}$
and bias vector $\beta \in {\Bbb R}^{n_{h}}$, where $n_{h}$ denotes 
the number of hidden units. Let us denote the unknown parameter vector
as $\theta = [w; V(:); \beta]$ which stacks all the weights into one single
vector. Since the introduction of the backpropagation algorithm
for training of neural networks [Rumelhart {\em et al.}, 1986; Werbos, 1990], 
important insights have been 
obtained about regularization (weight decay) of the cost function 
in order to have a good generalization ability in view of the 
bias-variance trade-off [Bishop, 1995; MacKay, 1992; Suykens \& Vandewalle, 1998]. 
It is well-known that instead of optimizing the cost function
\be
\label{cost}
\min_{\theta} J(\theta) = \frac{1}{2} \sum_{k=1}^{N} [d_{k} - y_{k}(\theta)]^{2}
\ee
where $d_{k}$ are the desired target outputs (with given data set 
$\{ u_{k}, d_{k} \}_{k=1}^{N}$ of $N$ training data), one 
better considers the following cost function with regularization
\be
\label{costreg}
\min_{\theta} J^{reg}(\theta; \mu, \zeta) = 
\zeta \frac{1}{2} \sum_{k=1}^{N} [d_{k} - y_{k}(\theta)]^{2} + \mu \frac{1}{2} \theta^{T} \theta 
\ee
where $\mu, \zeta$ are additional hyperparameters which can 
be automatically tuned within the framework of Bayesian learning.
The second term is related to a prior distribution on the weights
which typically expresses that the initial weights are chosen small.
The first term is associated with the likelihood.
When networks are overparametrized (too many hidden units), 
regularization enables to work implicitly with a number of effective parameters that is less
than the actual number of interconnection weights such that overfitting
can be avoided.

Let us now consider a CLM by taking as state vectors ${\bf x}^{(i)}$ 
($i=1,...,q$) the unknown parameter vectors $\theta^{(i)}$.
Hence, the CLM consists in fact of coupled backpropagation processes
with weight vector synchronization.
In Fig.~2 an example of estimating a sine function from given noisy
data is given where an overparametrized MLP is 
taken. Simulation results show that by CLMs one is able to 
obtain a similar smooth solution as from (\ref{costreg})
with Bayesian learning but by optimizing the original cost function
(\ref{cost}) without regularization. In order to obtain
this intelligent solution (i.e.~a good generalization performance),
the initial distribution of ${\bf x}^{(i)}(0)$ ($i=1,...,q$) (at time 0) plays
an important role, similar with the choice of $\mu$ in (\ref{costreg}).
Consider the following distribution of the initial CLM state
\be
\label{initial}
p({\bf x}(0)) \propto 
\exp[- \frac{1}{2 \sigma^{2}} {\bf x}(0)^{T} {\bf x}(0)]
\ee
which assumes that the states of the local minimizers ${\bf x}^{(i)}(0)$
are normally distributed. The value of $\sigma$ then plays the role
of a bifurcation hyperparameter. For $\sigma$ chosen very large
a solution with bad generalization is obtained, while for $\sigma$ small
(small initial weights) a good generalization performance is obtained.
Hence, CLMs can avoid the overfitting phenomenon {\em without} the use 
of a regularization in the sum squared error cost function.
In order to obtain the results with good generalization in Fig.2,
the following CLM tuning parameters were taken:
$q=20$, $\sigma =0.1$, $\alpha = 10^5$,
$\underline{\gamma} = 10^6$, $\overline{\gamma} = 10^7$,
$\underline{\eta} = 10^{-2}$, $\overline{\eta} = 10^{10}$,
$U^{*} = 0$ and 20\% of the local minimizers is re-numbered
after $c=5$ with $\Delta T = 10^{-9}$.


\section{Optimization of Lennard-Jones Clusters}

In this Section we discuss the application of CLMs to 
the optimization of Lennard-Jones clusters.
In predicting the three-dimensional structure of proteins from 
amino acid sequences, potential energy surface (PES) minimization
is often related to the native structure of the protein. Optimization
of Lennard-Jones (LJ) clusters is considered to be an 
important benchmark problem at this point [\v{S}ali {\em et al.}, 1994; Neumaier, 1997;
Wales \& Scheraga, 1999].
Recently, detailed studies have been reported about optimization
of (LJ)$_{N}$ with $N \leq 150$ by means of basin-hopping techniques
in comparison with many other approaches [Wales \& Doye, 1997; 
Wales {\em et al.}, 1998; Wales \& Scheraga, 1999]. 
The cost function in this case is given by 
\be
\label{LJ}
U_{LJ} =  4 \sum_{i<j} (\frac{1}{r_{ij}^{12}} 
- \frac{1}{r_{ij}^{6}})
\ee
where $r_{ij}$ the Euclidean distance between atom $i$ and $j$ ($j=1,...,N$).
(LJ)$_{38}$ which possesses a {\em double-funnel} energy landscape and is known
to be an interesting test-case.

Fig.~3 shows the result of a CLM run applied to (LJ)$_{38}$.
The CLM tuning parameters in this case are
$q=50$, $\alpha = 10^5$,
$\underline{\gamma} = 10^6$, $\overline{\gamma} = 10^7$,
$\underline{\eta} = 10^{-2}$, $\overline{\eta} = 10^{6}$,
$U^{*} = 0$ ($U_{LJ}^{\delta} = U_{LJ} + \delta$ with $\delta = 200$ is optimized
in order to ensure a positive cost function everywhere) and 10\% of the 
local minimizers is re-numbered after $c=5$ with $\Delta T = 10^{-8}$.
The CLM energy {\em bundle} caused by state synchronization of the 
local minimizers is clearly visible. 
Like in the application of CLMs to the training of neural networks,
the choice of $\sigma$ in $p({\bf x}(0))$ plays an important role. 
In Fig.~3 the differences are illustrated for $\sigma =0.1$ versus $\sigma =1$.
The choice of this prior implicitly corresponds in fact to an 
additional confining potential energy term, which has been explicitly considered 
in effective potential minimization methods [Schelstraete \& Verschelde, 1997] 
(in a different form). 
However, the effective potential minimization method has problems with detecting 
e.g.~the global minimum of (LJ)$_{8,9}$, which on the other hand are easily found by CLMs. 
In the example of (LJ)$_{38}$ a classical local optimization technique
(quasi-Newton and scaled conjugate gradient for large scale problems) has been used 
for post-processing and fine-tuning (about 10-20 additional iterations) of  
the CLM results. The variety of generated solutions includes the global minimum -173.9284.

In the application of CLMs to larger clusters, first a shifted 
version of the problem $U_{LJ}^{shift} =  4 \sum_{i<j} [\frac{1}{(r_{ij} + \mu)^{2 \nu}} 
- \frac{1}{(r_{ij} + \mu)^{\nu}}]$ has been minimized in order to avoid 
ill-conditioning in the LP problem and excessively large values in the
initial CLM evolution. For (LJ)$_{150}$ $\mu = 0.1$, $\nu =3$ has been chosen. 
These results have been taken then as initialization for 
optimization of $U_{LJ}$ and are, eventually, post-processed by a classical
local optimization method.
The CLM tuning parameters in the (LJ)$_{150}$ case are
$q=100$, $\sigma =0.1$, $\alpha = 10^5$, $U^{*} = 0$, 
$U_{LJ}^{shift,\delta} = U_{LJ}^{shift} + \delta$ with $\delta = 3000$
and $U_{LJ}^{\delta} = U_{LJ} + \delta$ with $\delta = 1000$.
The other tuning parameters for (LJ)$_{150}$ are the same
as for the (LJ)$_{38}$ example. 
This CLM run yields a PES value of -872.91, while the best known
minimum is -893.31 in this case, obtained by counting nearest-neighbour
interactions for icosahedral packing schemes (Northby). However,
like simulated annealing and basin-hopping approaches, our focus
is here on unbiased methods which are generally applicable.

One can argue that CLMs are providing good solutions {\em given}
the user-defined initial distribution (prior).
In this sense, CLM results can potentially be further improved 
by exploiting more informative prior knowledge about the problem.
The simulation results show that this becomes more important for larger clusters.
Simulations have been done on a Sun Ultra 2 workstation in Matlab with 
cmex implementation of the PES evaluation and gradient calculations.
CLM running times are roughly comparable with $q$ times applying a standard gradient
based local optimization methods (conjugate gradient).


\section{Conclusions}

We have introduced a new optimization method of coupled local minimizers.
This is done by considering the average energy cost of the 
total ensemble subject to pairwise synchronization constraints
that are asymptotically reached. The resulting 
continuous-time optimization algorithm from Lagrange programming
networks is an array of coupled nonlinear cells or CNN.
For problems of MLP training in supervised learning, we have 
shown that CLMs are able to produce intelligent solutions,
in the sense that one can obtain a good generalization ability
without taking a regularization term in the cost function.
The obtained insights are consistent with Bayesian learning 
methods for MLPs. We have also shown how one can let 
the ensemble of local minimizers cooperate in an optimal 
way. This is done by solving additional linear programming 
problems. The CLM method has been successfully applied to 
the optimization of Lennard-Jones clusters which is an
important benchmark problem in the area of protein folding.
We expect that the proposed CLM methodology may offer new insights for 
the application of continuous-time optimization algorithms to NP
complete problems in general. It also emphasizes the importance
of the role played by the initial state distribution
of the local minimizers that form the CLM ensemble. \\ \\ \\


\noindent
{\normalsize \bf Acknowledgements.} {\small  This research work was carried out 
      at the ESAT laboratory
      and the Interdisciplinary Center of Neural Networks ICNN
      of the Katholieke Universiteit Leuven, in the framework of the 
      FWO projects {\em Learning and Optimization: an 
      Interdisciplinary Approach} and {\em Collective Behaviour and Optimization:
      an Interdisciplinary Approach}, the Belgian
      Program on Interuniversity Poles of Attraction,
      initiated by the Belgian State, Prime Minister's Office for
      Science, Technology and Culture (IUAP P4-02 \& IUAP P4-24) and 
      the Concerted Action Project MEFISTO of the Flemish Community.
      Johan Suykens is a postdoctoral researcher with the 
      National Fund for Scientific Research FWO - Flanders.
}

\newpage

\noindent
{\Large \bf References}

\begin{description}
\item[]
Arbib, M.A. (Ed.) [1995]
{\em The Handbook of Brain Theory and Neural Networks}
(MIT Press, Cambridge MA).

\item[]
Bertsekas, D.P. [1996]
{\em Constrained Optimization and Lagrange Multiplier Methods}
(Athena Scientific).

\item[]
Bishop, C.M. [1995]
{\em Neural networks for pattern recognition}
(Oxford University Press).

\item[]
Chen, G. \& Dong, X. [1998]
{\em From Chaos to Order - Perspectives,
Methodologies, and Applications} 
(World Scientific Pub.~Co., Singapore).

\item[]
Chua, L.O. \& Roska, T. [1993] 
``The CNN Paradigm,''
{\em IEEE Trans.~Circuits and Systems-I}, {\bf 40}(3), 147-156.

\item[]
Chua, L.O., Hasler, M., Moschytz, G.S. \& Neirynck, J. [1995]
``Autonomous Cellular Neural Networks: a Unified Paradigm for
Pattern Formation and Active Wave Propagation,''
{\em IEEE Trans.~Circuits and Systems-I}, {\bf 42}(10), 559-577.

\item[]
Chua, L.O. [1998]
{\em CNN: a Paradigm for Complexity}
(World Scientific Pub.~Co., Signapore).

\item[]
Cichocki, A. \& Unbehauen, R. [1994]
{\em Neural Networks for Optimization and Signal Processing}
(Wiley, Chichester).

\item[]
Fletcher, R. [1987]
{\em Practical Methods of Optimization}
(Wiley, Chichester).

\item[]
Gill, P.E., Murray, W. \& Wright, M.H. [1981]
{\em Practical Optimization}
(Academic Press, London).

\item[]
Goldberg, D.E. [1989]
{\em Genetic Algorithms in Search, Optimization and Machine Learning}
(Addison-Wesley, Reading Mass.).

\item[]
Kirkpatrick, S., Gelatt, C.D. \& Vecchi, M. [1983]
``Optimization by simulated annealing,''
{\em Science} {\bf 220}, 621-680.

\item[]
MacKay, D.J.C. [1992]
``Bayesian Interpolation,''
{\em Neural Computation} {\bf 4}(3), 415-447.

\item[]
Neumaier, A. [1997]
``Molecular modeling of proteins and mathematical prediction of
protein structure,''
{\em SIAM Rev.} {\bf 39}, 407-460. 

\item[]
Pecora, L.M. \& Carroll, T.L. [1990]
``Synchronization in Chaotic Systems,''
{\em Phys.~Rev.~Lett.} {\bf 64}, 821-824.

\item[]
Poggio, T. \& Shelton, C. [1999]
``Machine Learning, Machine Vision and the Brain,''
{\em AI Magazine} {\bf 20}(3), 37-55.

\item[]
Rumelhart, D.E., Hinton, G.E. \& Williams, R.J. [1986]
``Learning representations by back-propagating errors,''
{\em Nature} {\bf 323}, 533-536.

\item[]
\v{S}ali, A., Shakhnovich, E. \& Karplus, M. [1994]
``How does a protein fold?,''
{\em Nature} {\bf 369}, 248-251.

\item[]
Schelstraete, S. \& Verschelde, H. [1997]
``Finding minimum-energy configurations of Lennard-Jones clusters
using an effective potential,''
{\em J. Phys. Chem. A} {\bf 101}, 310-315.

\item[]
Styblinski, M.A. \& Tang, T.-S. [1990]
``Experiments in nonconvex optimization: stochastic approximation
with function smoothing and simulated annealing,''
{\em Neural Networks} {\bf 3}, 467-483.

\item[]
Suykens, J.A.K., Vandewalle, J.P.L. \& De Moor, B.L.R. [1996]
{\em Artificial Neural Networks for Modelling and Control of
Non-Linear systems}, 
(Kluwer Academic Publishers, Boston).

\item[]
Suykens, J.A.K., Curran, P.F., Vandewalle, J. \& Chua, L.O. [1997]
``Robust nonlinear H$_{\infty}$ synchronization of chaotic Lur'e systems,''
{\em IEEE Trans.~Circuits and Systems-I}
{\bf 44}, 891-904.

\item[]  
Suykens, J.A.K. \& Vandewalle, J. (Eds.) [1998]
{\em Nonlinear Modeling: advanced black-box techniques}
(Kluwer Academic Publishers, Boston).

\item[]
Suykens, J.A.K., Yang, T. \& Chua, L.O. [1998]
``Impulsive synchronization of chaotic Lur'e systems 
by measurement feedback,''
{\em Int.~J.~Bifurcation and Chaos}
{\bf 8}, 1371-1381.

\item[]
Suykens, J.A.K. \& Vandewalle, J. [2000]
``Chaos synchronization: a Lagrange programming network approach,''
{\em Int.~J.~Bifurcations and Chaos
(special issue on chaos control and synchronization)}, 
{\bf 10}(4), 797-810.

\item[]
Wales, D.J. \& Doye, J.P.K. [1997]
``Global optimization by basin-hopping and the lowest energy structures
of Lennard-Jones clusters containing up to 110 atoms,''
{\em J.~Phys.~Chem.~A} {\bf 101}, 5111-5116.

\item[]
Wales, D.J., Miller, M.A. \& Walsh, T.R. [1998]
``Archetypical energy landscapes,''
{\em Nature} {\bf 394}, 758-760.

\item[]
Wales, D.J. \& Scheraga, H.A. [1999]
``Global optimization of clusters, crystals, and biomolecules,''
{\em Science} {\bf 285}, 1368-1372.

\item[]
Watts, D. J. \& Strogatz, S.H. [1998]
``Collective dynamics of 'small-world' networks,''
{\em Nature} {\bf 393}, 440-442. 

\item[]
Wu, C.W. \& Chua, L.O. [1994]
``A unified framework for synchronization and control 
of dynamical systems,''
{\em Int.~J.~Bifurcation and Chaos}
{\bf 4}, 979-989.

\item[]
Werbos, P. [1990]
``Backpropagation through time: what it does and how to do it,''
{\em Proceedings of the IEEE} {\bf 78}(10), 1150-1560.

\item[]
Zhang, S. \& Constantinides, A.G., [1992]
``Lagrange programming neural networks,''
{\em IEEE Trans.~Circuits and Systems-II}
{\bf 39}, 441-452.

\end{description}


\newpage

\noindent
{\large \bf Captions of Figures} \\

\noindent
{\bf Fig.~1} \, Illustration of the basic state synchronization
mechanism in coupled local minimizers (CLM):
({\bf A}) double well cost function $U(x) = x^{4} - 16 x^{2} + 5 x + 100$ 
with global minimum located at $x = -2.90$; 
({\bf B}) CLM with 
$\dot{x} = - \nabla U(x) - (x-z) - \lambda$,
$\dot{z} = - \nabla U(z) + (x-z) + \lambda$,
$\dot{\lambda} = x - z$,
where $x, z \in {\Bbb R}$ (red and blue, respectively) 
are the two particle states with costate $\lambda \in {\Bbb R}$ (green).
The CLM corresponds to the Lagrange programming network
with cost function $U(x) + U(z)$ subject to $x = z$.
Except when both initial states $x(0), z(0)$ are positive, 
the global minimum is always reached by this CLM.
Similar results are obtained for other double well problems with 
broad and narrow minima.
\\ \\
\noindent
{\bf Fig.~2} \, Supervised learning of multilayer perceptrons
on a benchmark problem of training a sinusoidal function (green) given 
20 training data points $\{ u_{k}, d_{k} \}$
(blue circles) corrupted with Gaussian noise (zero mean,
standard dev.~0.4) using a multilayer perceptron with one hidden layer 
consisting of 10 hidden units. Given the fact that the MLP consists
of 30 unknown parameters, overfitting will occur with 
standard training methods, unless regularization or early stopping is applied.
In ({\bf A}) results from standard neural network training methods are shown:
scaled conjugate gradient without early stopping leading to overfitting (red);
best result by Bayesian learning with regularization of the cost function
and automatic hyperparameter selection, leading to good generalization (blue).
({\bf B}) shows that a CLM which optimizes a sum squared error 
on the training data without regularization of the cost function
can produce a result (blue)
which is comparable to the quality of the Bayesian learning solution in 
({\bf A}),
provided the $\sigma$ value in $p({\bf x}(0)) \propto 
\exp(- \frac{1}{2 \sigma^{2}} {\bf x}(0)^{T} {\bf x}(0))$ is well
chosen (here $\sigma = 0.1$). ({\bf C}) shows
a result from a CLM for a bad choice of $\sigma = 5$. 
The large variance among the resulting local minimizers of the CLM is 
clearly visible, in comparison with ({\bf B}) where the variance is small
with good generalization. In ({\bf D}) the CLM evolution with $q=20$ of the sum 
squared error cost function is shown during optimization, 
related to the results of ({\bf B}). \\ \\
\noindent
{\bf Fig.~3} \, CLM application to potential energy surface minimization
of Lennard-Jones clusters: ({\bf A}) CLM evolution of $U_{LJ}^{\delta}$ 
for $q=50$ on (LJ)$_{38}$;
({\bf B}) Importance of the choice of the initial distribution:
$\sigma = 1$ (red), $\sigma = 0.1$ (blue) with post-processing by
means of a quasi-Newton method resulting into 
$\sigma = 1$ (magenta), $\sigma = 0.1$ (green), respectively.
Shown are the mean (dashed line), min and max (solid lines)
over 7 CLM runs after sorting the energy values. The variety of solutions
includes the global minimum -173.9284, visualized in ({\bf C}).
In ({\bf D}) a CLM result with $q=100$ is 
shown for a larger cluster (LJ)$_{150}$.

\newpage

\centerline{
{\bf A}
\psfig{figure=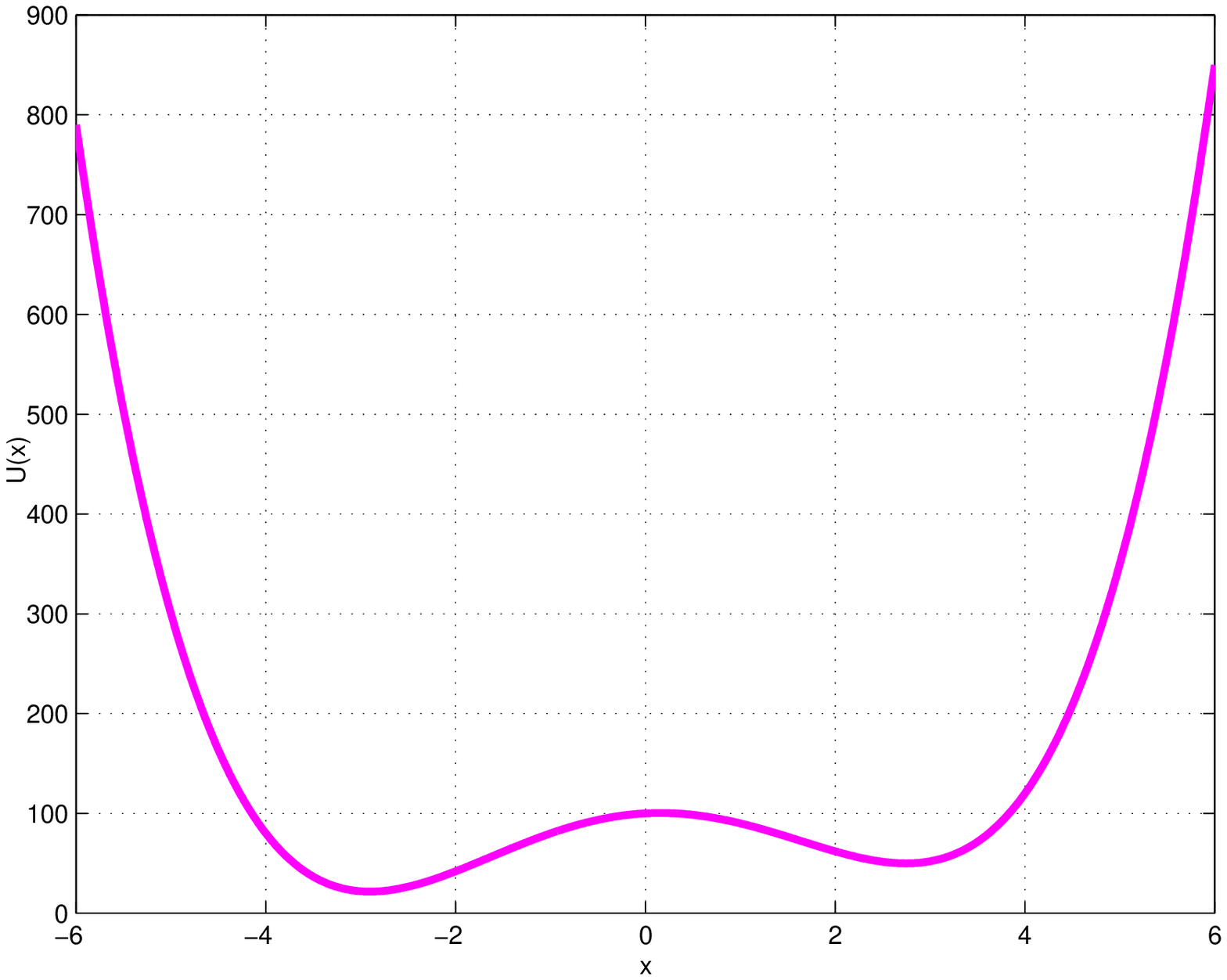,width=11cm}
}
\vspace*{10mm}
\centerline{
{\bf B}
\psfig{figure=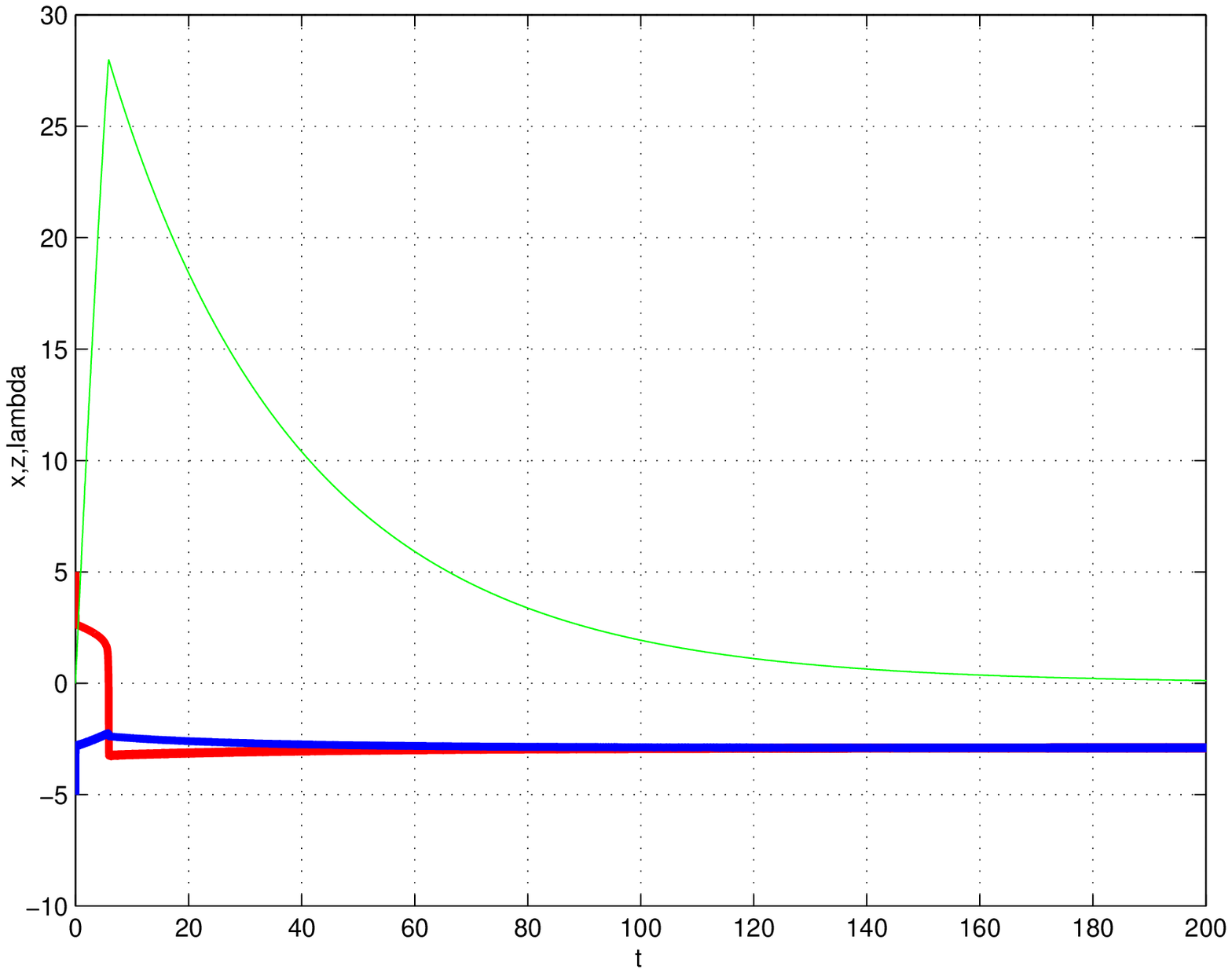,width=11cm}
}
\begin{quote}
{\bf Fig.~1}
\end{quote}

\newpage

\centerline{
{\bf A}
\psfig{figure=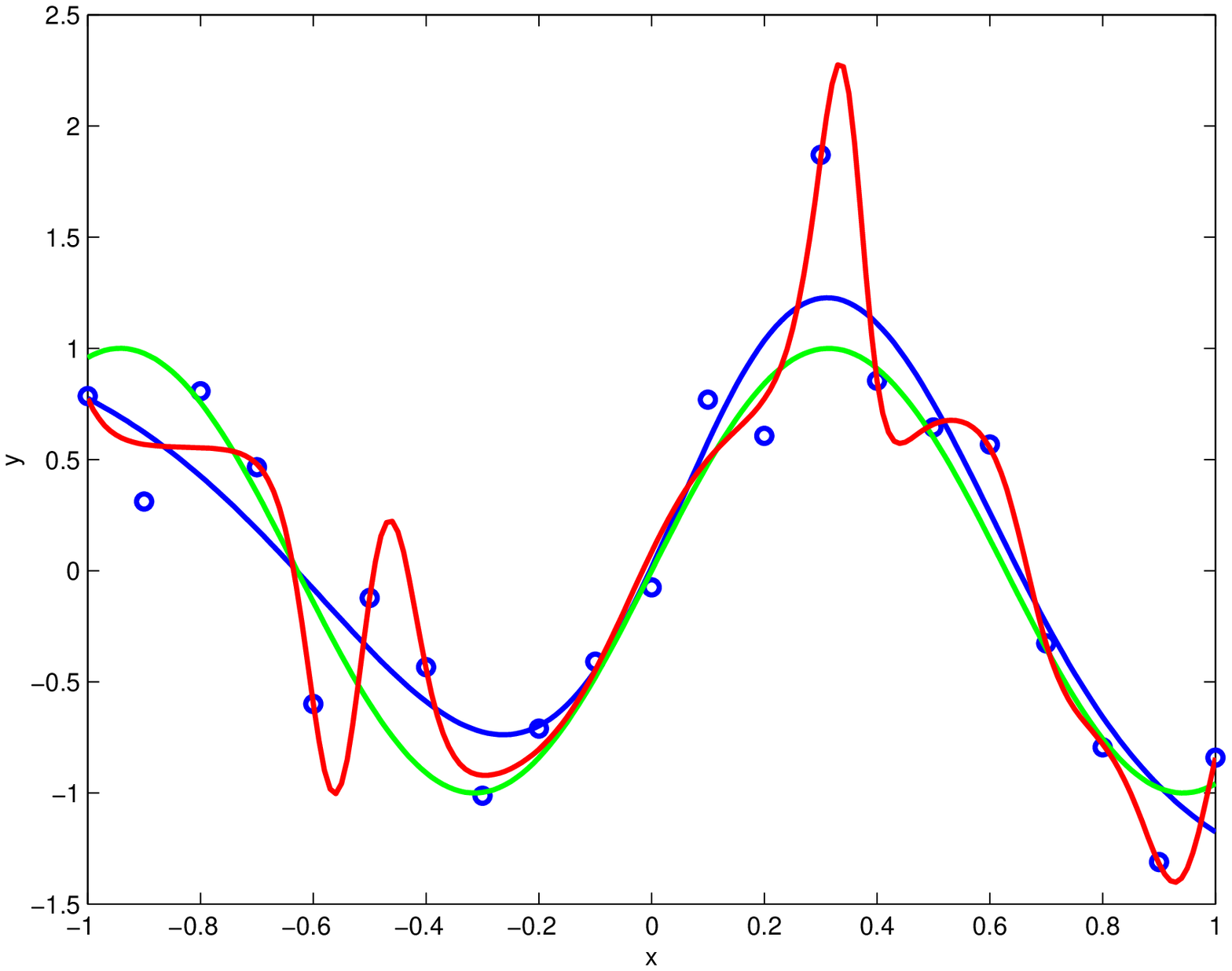,width=11cm}
}
\vspace*{10mm}
\centerline{
{\bf B}
\psfig{figure=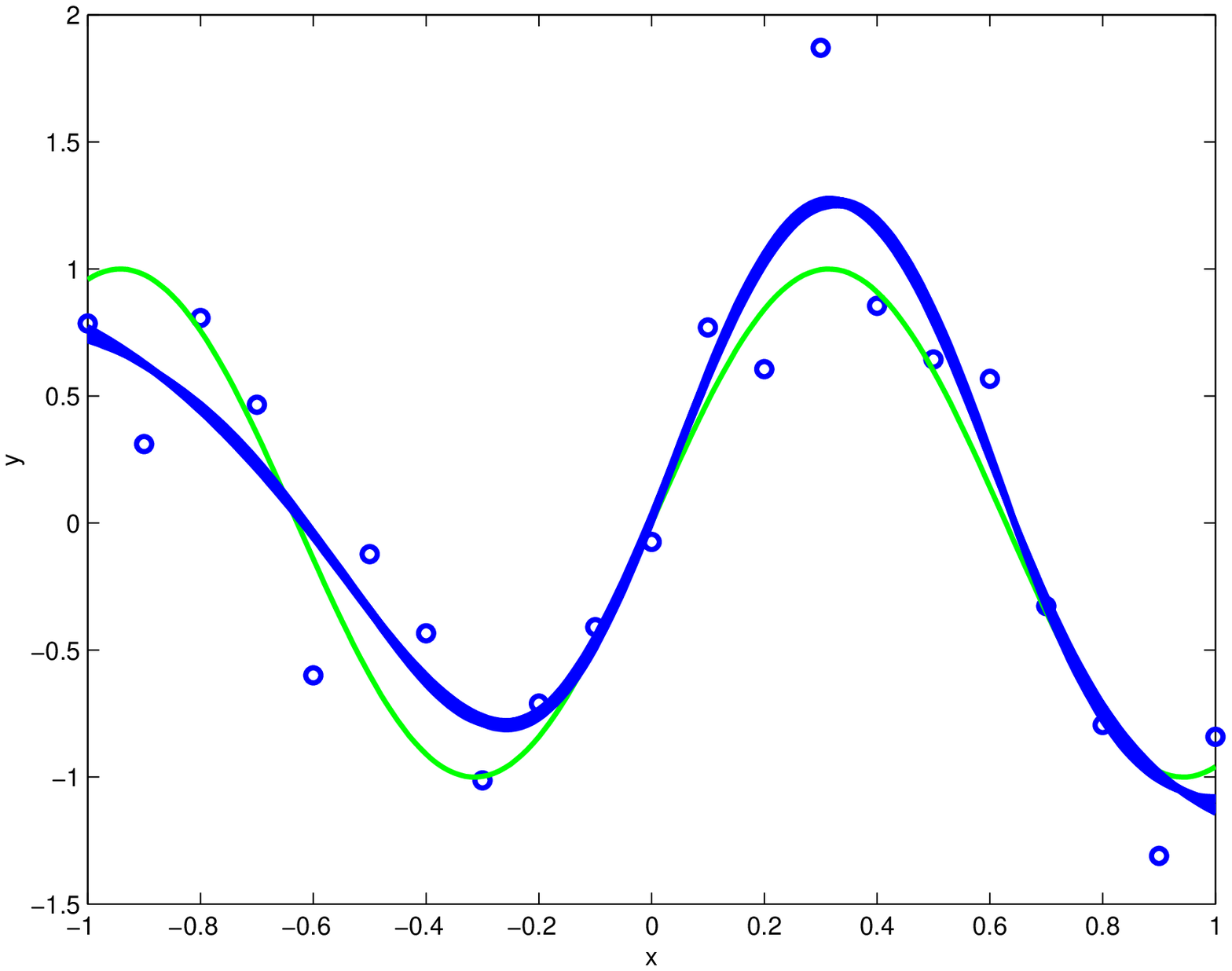,width=11cm}
}

\newpage

\centerline{
{\bf C}
\psfig{figure=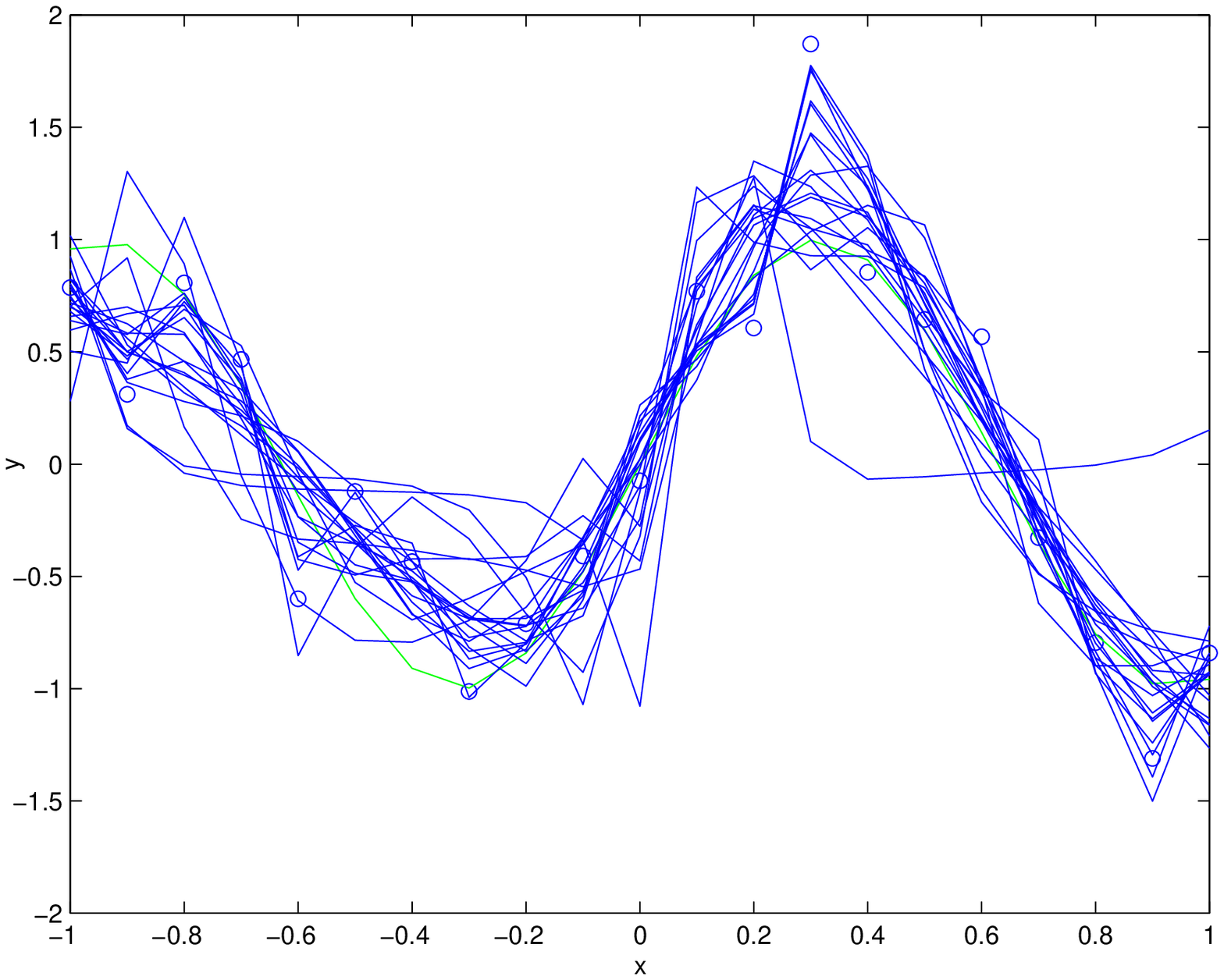,width=11cm}
}


\vspace*{10mm}
\centerline{
{\bf D}
\psfig{figure=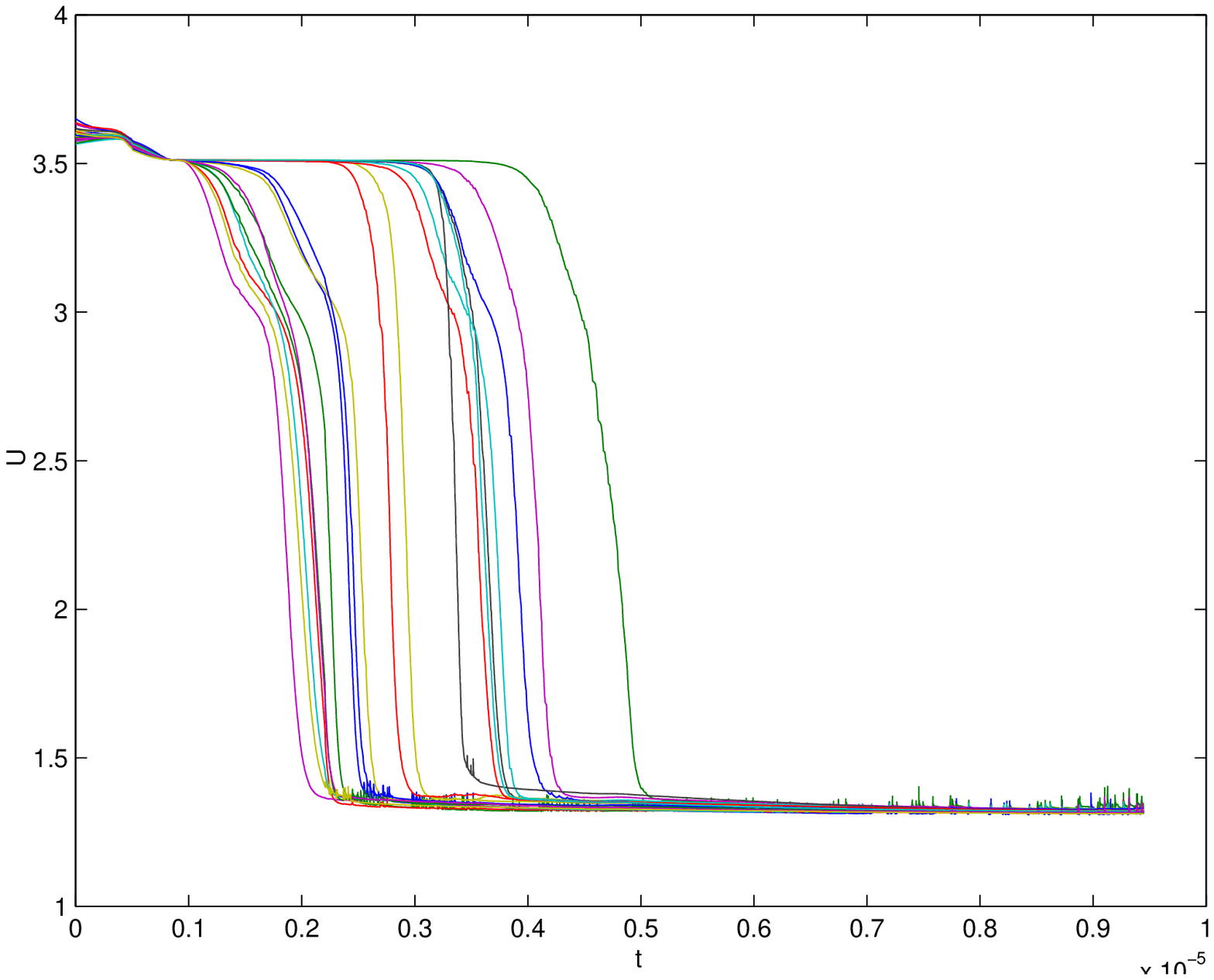,width=11cm}
}

\begin{quote}
{\bf Fig.~2}
\end{quote}

\newpage

\centerline{
{\bf A}
\psfig{figure=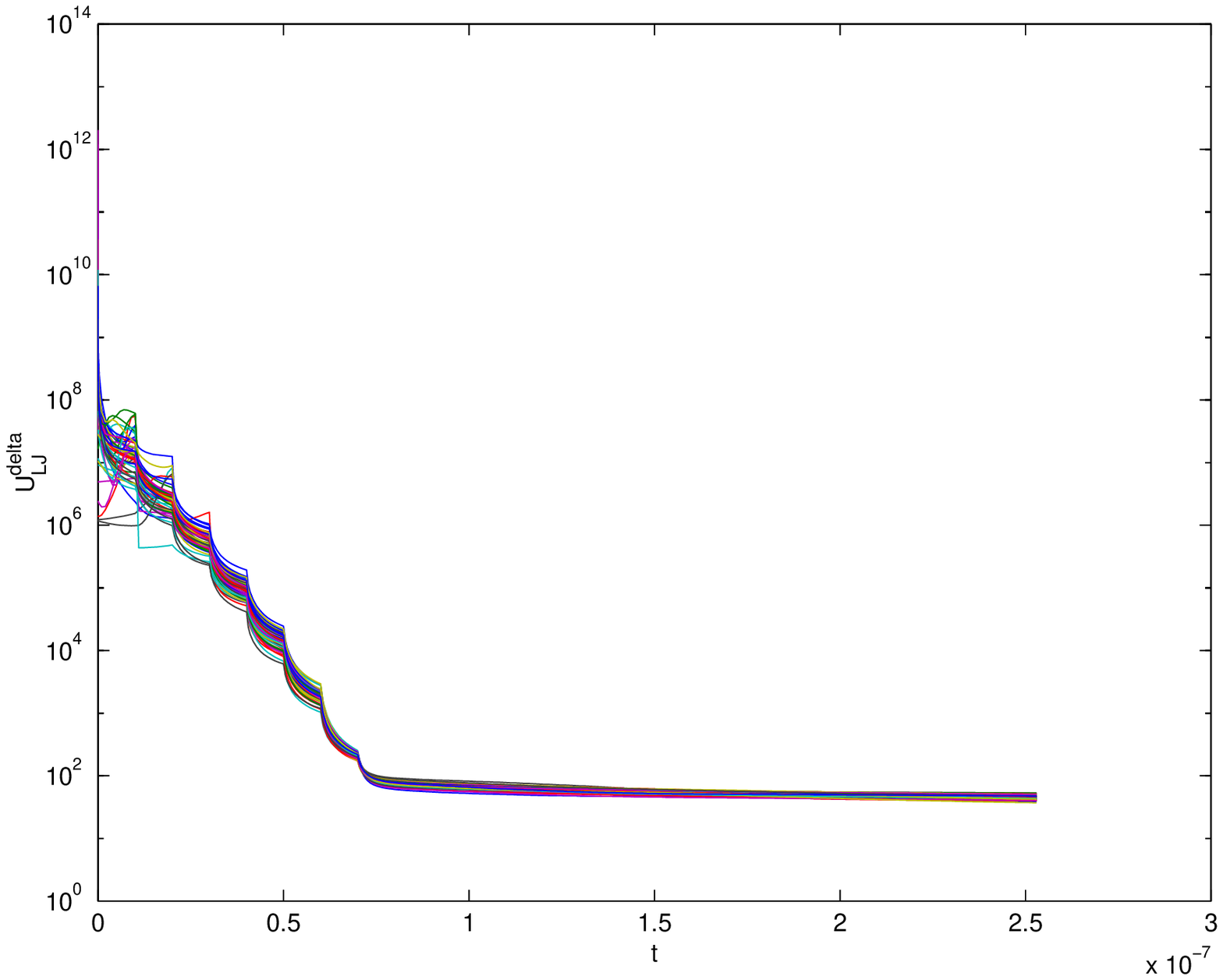,width=11cm}
}
\vspace*{10mm}
\centerline{
{\bf B}
\psfig{figure=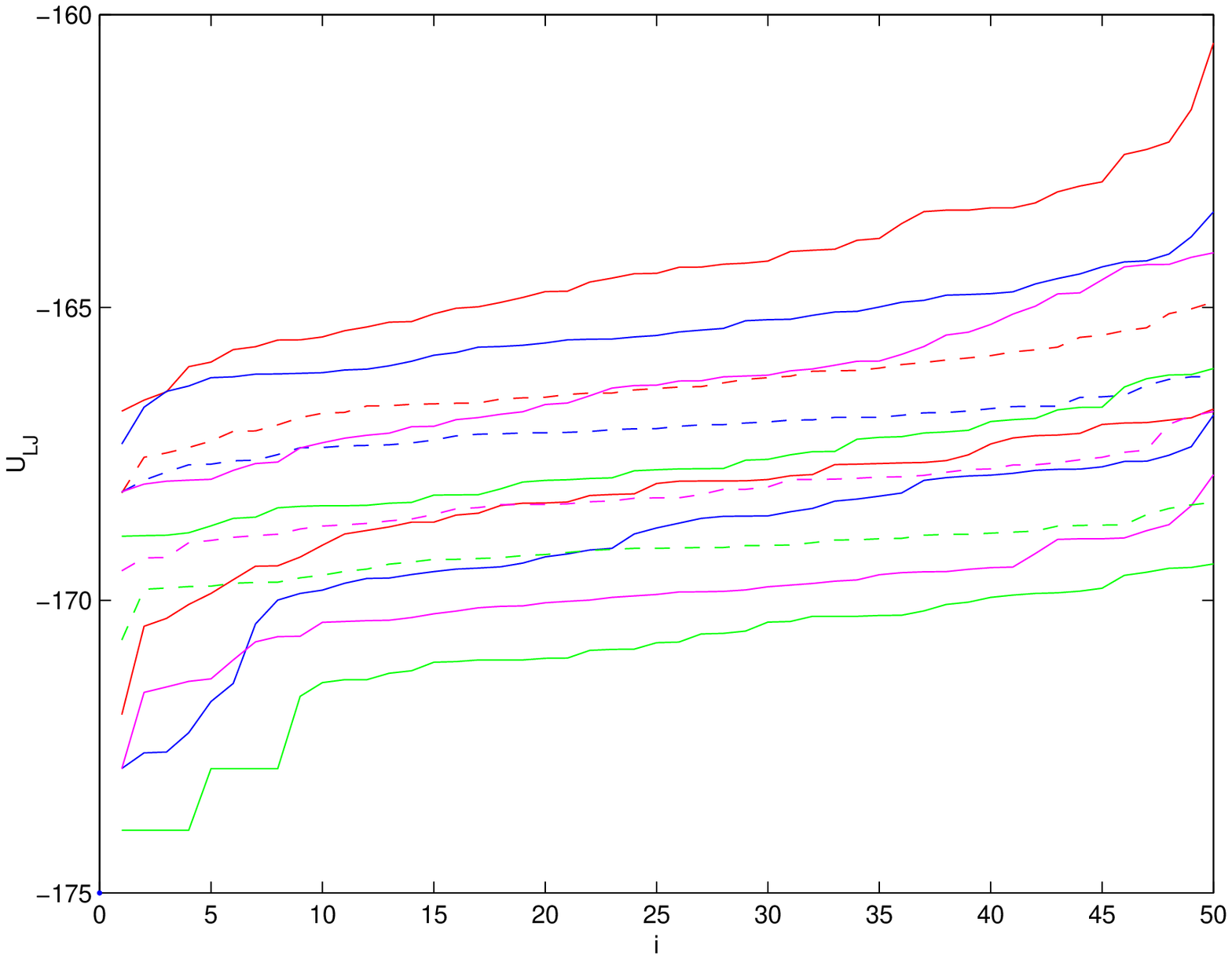,width=11cm}
}

\newpage


\centerline{
{\bf C}
\psfig{figure=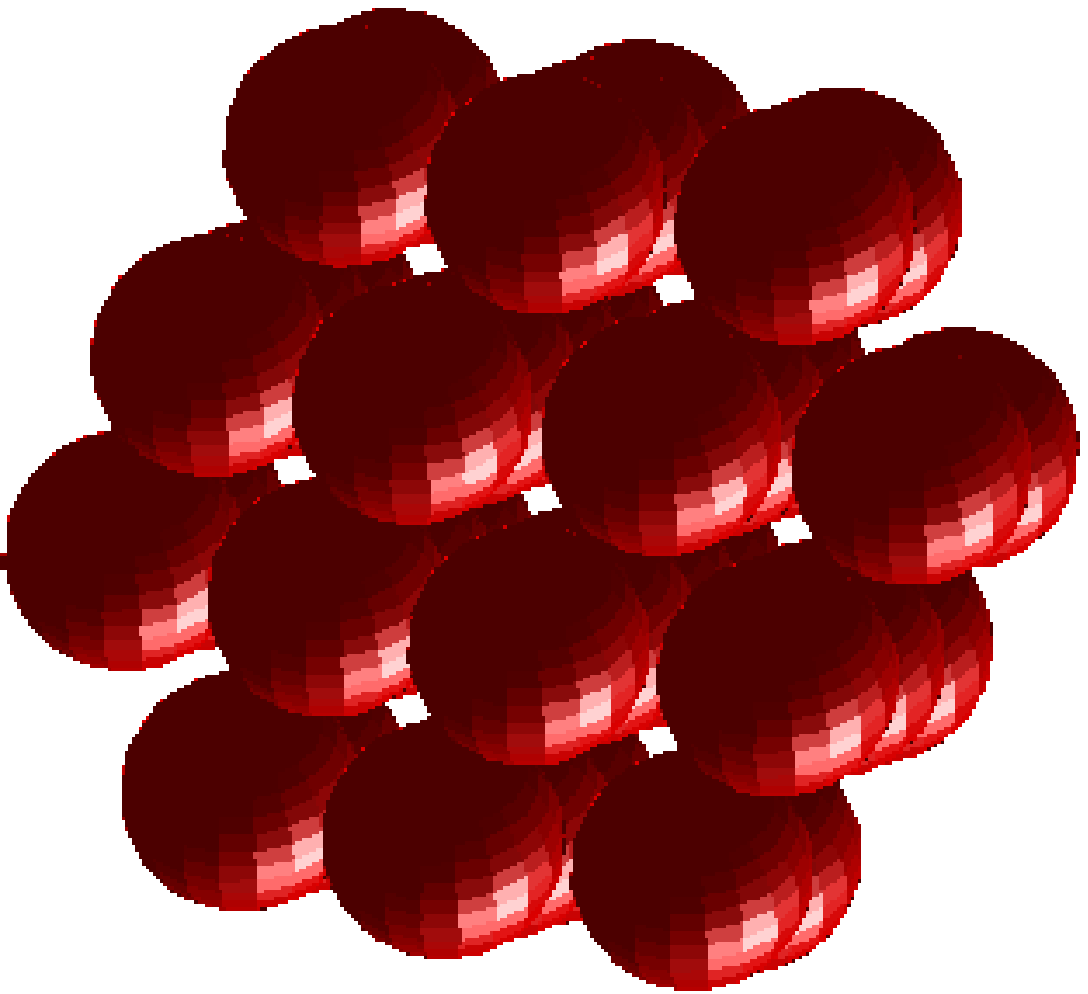,width=8cm}
}
\vspace*{10mm}
\centerline{
{\bf D}
\psfig{figure=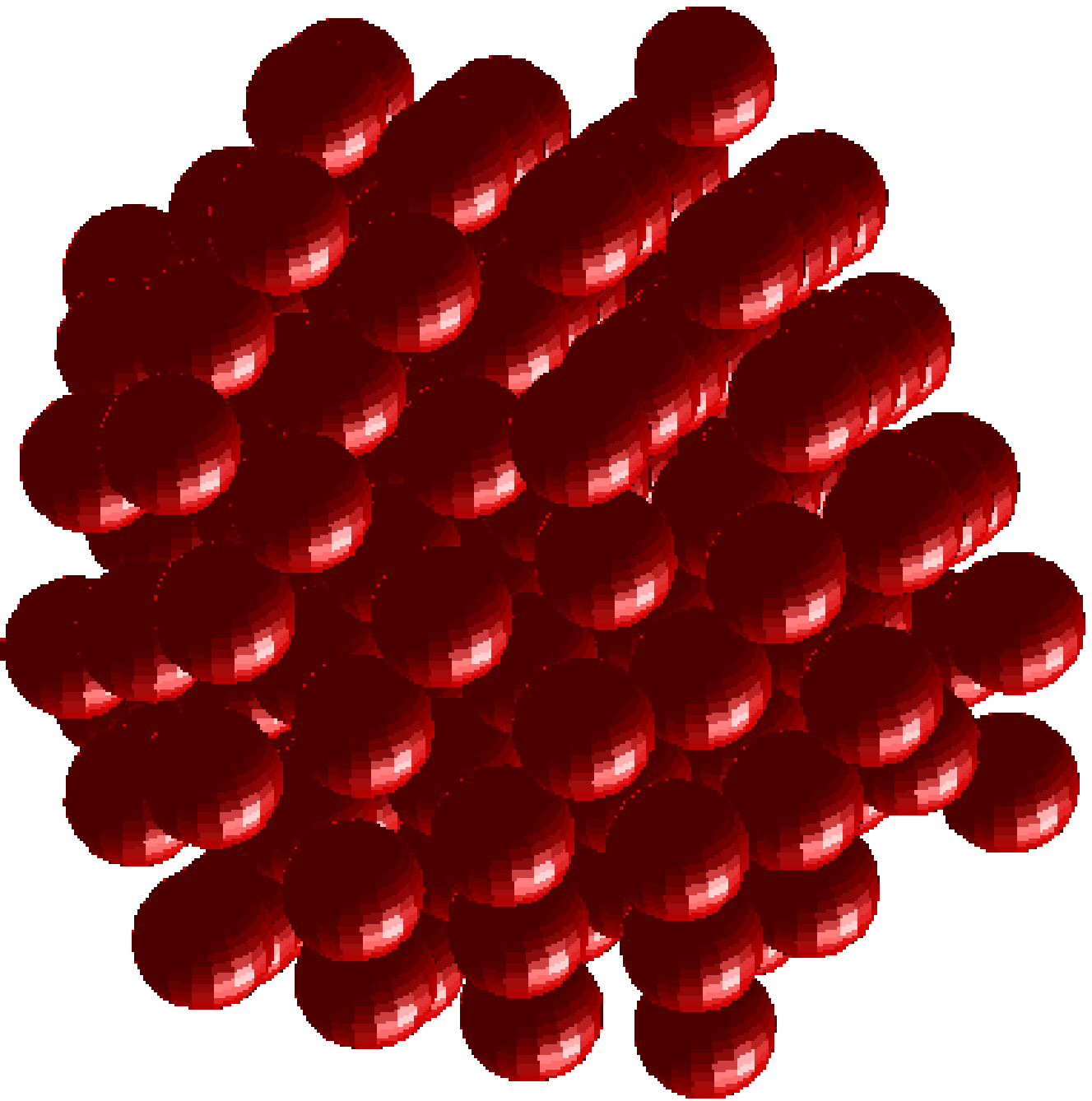,width=9cm}
}
\begin{quote}
{\bf Fig.~3}
\end{quote}

\end{document}